\documentclass[runningheads]{llncs}
\usepackage[T1]{fontenc}
\usepackage{graphicx}
\usepackage{booktabs}
\usepackage{multirow}
\usepackage{amsmath}
\usepackage{caption}
\newsavebox{\topbox}

\makeatletter
\usepackage{cleveref}
\usepackage{marvosym}
\begin{document}
\title{Seeing Through Multiple Views:
Parameter-Efficient Fine-Tuning via Selective
Neurons for Consistent Radiology Report
Generation}
%
%
\author{Yucheng Chen\inst{1,2,3} \and
Jinjing Zhu\inst{4} \and
Yang Yu\inst{5} \and Yufei Shi\inst{1,2} \and Hane Naghshbandi\inst{1,2,3} \and Jinhua Liu \inst{1,2} \and Angela S. Koh \inst{6} \and Fang Fen \inst{5} \and Kian Eng Ong \inst{1} \and Si Yong Yeo \inst{1,2,3}\thanks{Corresponding author}}
\authorrunning{Y. Chen et al.}
%
\institute{MedVisAI Lab \and Lee Kong Chian School of Medicine, Nanyang Technological University, Singapore \and Centre of AI in Medicine, Singapore \and
The Chinese University of Hong Kong, Sha Tin, New Territories, Hong Kong
\and
Institute for Infocomm Research (I\textsuperscript{2}R), A*STAR, Singapore
\and
National Heart Centre Singapore (NHCS), Singapore
}
\maketitle              
\begin{abstract}
Recent years have seen substantial advances in radiology report generation (RRG), yet existing approaches predominantly adopt direct feature fusion when handling multi-view X-ray images. Such approaches overlook the potential clinical inconsistencies and inaccuracies arising when a single model processes different views, adversely impacting performance and clinical reliability. To this end, we introduce \textbf{View-PNDF} (\textbf{View}-specific \textbf{P}attern \textbf{N}euron \textbf{D}etection and \textbf{F}ine-tuning), a parameter-efficient framework that fosters view-consistent report generation from a neuronal perspective. Specifically, View-PNDF comprises: (i) a view-specific neuron detection module identifying neurons responsive to particular views, (ii) a verification module quantifying the existence of these neurons, and (iii) a selective fine-tuning strategy strengthening detected neurons while preserving view-agnostic representations. By updating only view-specific neurons, View-PNDF achieves consistent diagnoses across different views with reduced computational costs. Subsequently, we employ Large Language Models (LLMs) to consolidate the view-specific reports into a complete radiology report. Furthermore, we use traditional Natural Language Generation (NLG) metrics-based assessment on integrated reports for baseline comparison and employ LLM-based assessment (e.g., GPT-4o) on view-specific reports to capture clinical significance. Extensive experiments on two medical RRG benchmarks demonstrate that View-PNDF substantially improves view-specific chest X-ray report generation quality while maintaining robust general-view performance.

\end{abstract}
\section{Introduction}
\label{sec:intro}
Radiology report generation (RRG) aims to automatically generate free-text descriptions from radiological images such as X-rays~\cite{liang2024divide} and CT scans~\cite{hamamci2024ct2rep}. Current RRG approaches can be categorized into single-view and multi-view methods. Single-view models~\cite{chen2020generating,liu2023observation,nicolson2023improving,jin2024promptmrg,liu2024bootstrapping,huang2025damper} focus on generating reports from individual radiographic projections but suffer from inherent limitations in capturing comprehensive anatomical information. Multi-view approaches~\cite{chen2022cross,yang2023radiology,liu2025enhanced} address these limitations by incorporating information from multiple perspectives - such as frontal, lateral, posteroanterior (PA), and anteroposterior (AP) views - demonstrating empirical improvements over their single-view counterparts. This multi-view paradigm aligns with clinical practice, where radiologists routinely examine each imaging view individually and then integrate these per-view observations into a comprehensive report.

Multi-view medical imaging is clinically essential for identifying posterior abnormalities, localizing lesions, and detecting conditions obscured in single projections~\cite{ittyachen2017forgotten,yang2016cardiac,yeo2011level,loh2022explainable}. However, existing multi-view RRG methods~\cite{chen2022cross,miao2024evoke,yang2023radiology,liu2025enhanced} primarily employ fusion strategies to enhance feature representation, overlooking a critical issue: different views exhibit intrinsic variations that challenge automated systems. Such variations may lead to inconsistent interpretations across views, compromising diagnostic reliability and report quality. Current systems like InternVL-DA-Medical-4B~\cite{chen2024internvl} generate reports from frontal views that generally align with GT reports, while lateral view outputs often contain contradictory findings. Such inconsistencies pose significant clinical risks, particularly when frontal images are unavailable due to patient positioning constraints, trauma conditions, or equipment limitations, making lateral views the primary diagnostic input. In settings where frontal and lateral views are provided separately, inconsistent outputs may further lead to confusion, misinterpretation, and unreliable clinical decision-making. This raises a fundamental question: \textit{what underlying mechanisms cause these view-specific discrepancies, and how can we systematically address them?} Without understanding the root causes at the model's internal level, solutions may only address symptoms rather than the core problem. Inspired by~\cite{bau2017network,mu2020compositional}, this motivates neuron-level analysis to identify which computational components are responsible for view-dependent variations, enabling targeted interventions to systematically improve cross-view consistency.

In this paper, we present a comprehensive neuron-level analysis of RRG models, namely \textbf{View}-specific \textbf{P}attern \textbf{N}euron \textbf{D}etection and \textbf{F}ine-tuning (\textbf{View-PNDF}). Rather than retraining the entire model, we first introduce a View-specific Neuron Detection (\textbf{VND}) module to identify a compact set of pattern neurons responsible for view-specific inconsistency. We then apply a View-specific Neuron Verification (\textbf{VNV}) module to quantitatively evaluate the contribution of these detected neurons by selectively deactivating them and measuring the impact on view-specific report generation. Building on these insights, we further propose a neuron-level adaptation strategy, View-specific Neuron Fine-tuning (\textbf{VNF}), which focuses training exclusively on the identified pattern neurons. The targeted fine-tuning enables the model to generate consistent view-specific findings from any available view, while findings from multiple views can be further consolidated into an integrated radiology report using LLMs. For evaluation, we use standard Natural Language Generation (NLG) metrics to assess the quality of the integrated reports. However, evaluating view-specific reports presents unique challenges as conventional NLG metrics  fail to capture clinical validity: reports describing different anatomical views may receive similar scores even when they express contradictory clinical findings since NLG metrics measure textual overlap rather than medical semantics. To address this limitation, we employ a LLM–based evaluation method (e.g., GPT-4o) to assess the semantic fidelity and diagnostic accuracy of view-specific reports. By integrating view-aware methodology and appropriate evaluation strategies with RRG, our proposed framework offers a targeted and efficient solution for enhancing multi-view RRG performance. 

\section{Method}
\label{sec:method}

\begin{figure*}[t]
\centering
  \includegraphics[width=1\linewidth]{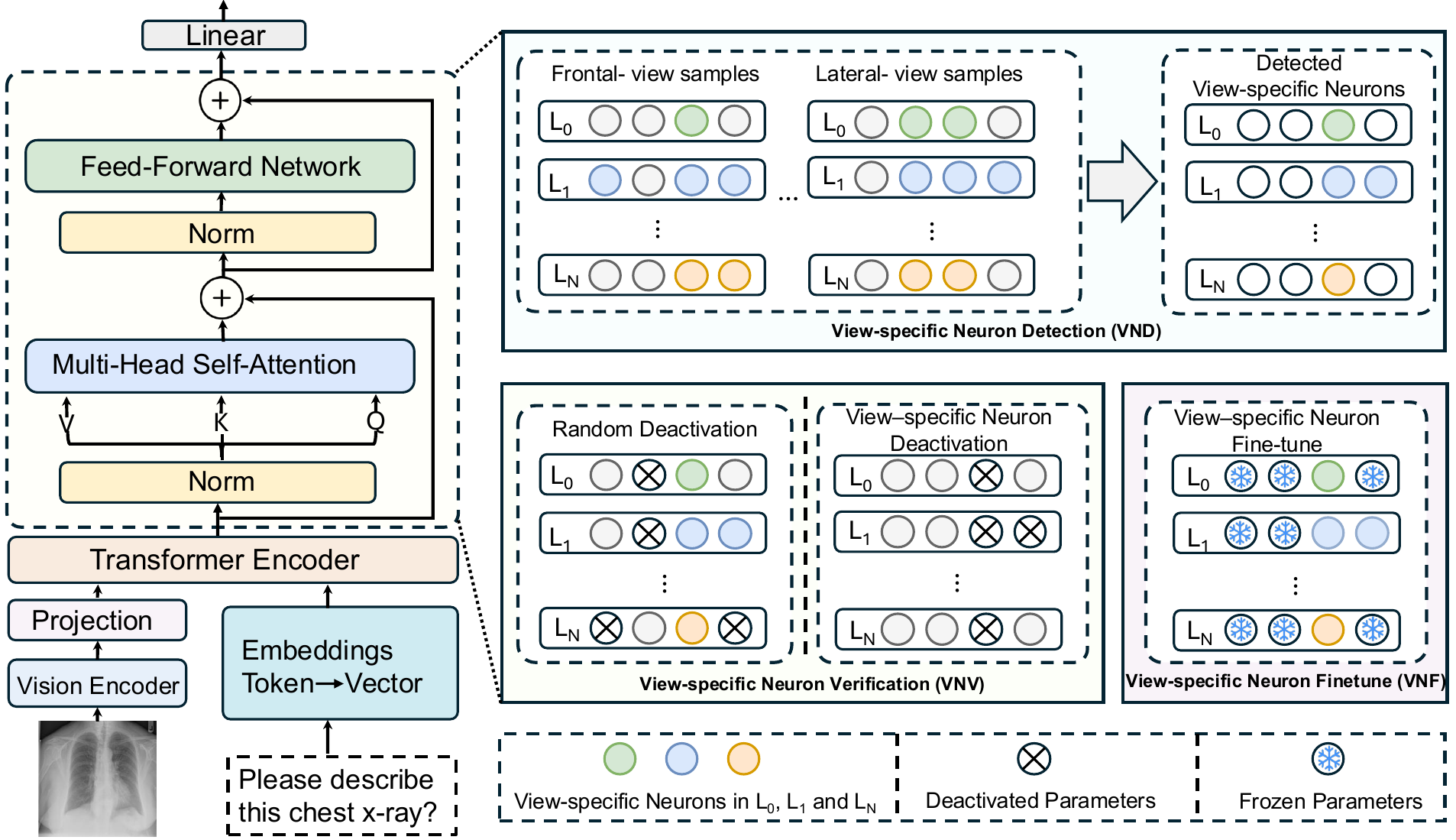}
\caption{The overall framework of View-PNDF, which consists of four components: the transformer structure in VLMs for RRG; View-specific Neuron Detection (VND) module identifies neurons associated with the specific views of chest X-rays; View-specific Neuron Verification (VNV) module verifies the detected pattern neurons; and View-specific Neuron Fine-tuning (VNF) module fine-tunes the detected view-specific neurons for RRG network.}
\label{fig:framework}  
\end{figure*}

In this section, we introduce \textbf{View}-specific \textbf{P}attern \textbf{N}euron \textbf{D}etection and \textbf{F}ine-tuning (\textbf{View-PNDF}), a framework designed to enhance Radiology Report Generation (RRG) by identifying and optimizing neurons responsible for processing specific X-ray views (e.g., frontal vs. lateral). As illustrated in Fig.~\ref{fig:framework}, our approach consists of three key components: View-specific Neuron Detection (\textbf{VND}), Neuron Verification (\textbf{VNV}), and Neuron Fine-tuning (\textbf{VNF}).

\subsection{View-specific Pattern Neuron Detector (VND)}
\label{sec:vnd}

We hypothesize that within the RRG model - specifically in the linear and self-attention layers of the Transformer decoder - distinct subsets of neurons are specialized for handling different X-ray views. We define these as \textit{View-specific Pattern Neurons}.

\noindent\textbf{Neuron Influence via Perturbation.} 
Ideally, the importance of a neuron can be measured by the impact of its deactivation on the model's output. Formally, let $\mathcal{RRG}(x, \theta)$ denote the generated embedding for a view-specific input sample $x$ (comprising visual features and prompt tokens) given model parameters $\theta$. The influence $I$ of the $k$-th neuron at the $l$-th layer, denoted as $n_l^k$, on a specific view pattern $v$ is defined as the perturbation effect:
\begin{equation}
I(n_l^k | x^v) = \left\| \mathcal{RRG}(x^v, \theta) - \mathcal{RRG}(x^v, \theta \setminus w_l^k) \right\|_2,
\label{equ:influence}
\end{equation}
where $\theta \setminus w_l^k$ represents the model weights with the specific neuron $n_l^k$ deactivated (zeroed out).

\noindent\textbf{Efficient Detection via Attribution Scores.}
Since calculating Eq.~\ref{equ:influence} for every neuron across the entire dataset is computationally prohibitive, we follow prior work~\cite{dai2021knowledge}, which suggests that neurons with high activation values are strong indicators of pattern importance. We therefore employ an efficient \textit{Attribution Score} as a proxy for detection. The attribution score for neuron $n_l^k$ given input $x^v$ is defined as the magnitude of its activation output:
\begin{equation}
\text{Attr}(n_l^k | x^v) = \left\| \sum_{t} f(x^v, w_l^k)_t \right\|_1,
\label{equ:attr}
\end{equation}
where $f(\cdot)$ computes the activation values (e.g., the output of a specific linear projection or attention head) across the sequence dimension $t$.

\noindent\textbf{Detection Strategy.}
To robustly identify neurons associated with a view $v$ (e.g., lateral), we utilize a set of detection samples $\mathcal{D}_{detect} = \{x^v_1, \dots, x^v_B\}$. A neuron is classified as a \textit{View-specific Pattern Neuron} if it consistently ranks among the most active neurons across all samples in the detection set. The set of pattern neurons $\mathcal{N}^v$ is formally defined as:
\begin{equation}
\mathcal{N}^v = \bigcap_{b=1}^{B} \left\{ n_l^k \mid \text{rank}(\text{Attr}(n_l^k | x^v_b)) \leq \gamma N \right\},
\label{equ:detection}
\end{equation}
where $\text{rank}(\cdot)$ sorts neurons in descending order of attribution score, $N$ is the total number of neurons in the layer, and $\gamma$ is a selection ratio (e.g., top 5\%). This intersection strategy ensures we select neurons that are consistently salient for the specific view, filtering out noise from instance-specific variations.

\subsection{Verification and Fine-tuning}

\textbf{View-specific Neuron Verification (VNV).}
To validate that the detected set $\mathcal{N}^v$ genuinely controls view-specific generation, we employ the perturbation metric from Eq.~\ref{equ:influence} on a held-out verification set. We compare the performance drop when deactivating the detected pattern neurons versus deactivating a random set of neurons of the same size. The existence of view-specific neurons is confirmed if:
\begin{equation}
\sum_{d=1}^{D} \mathcal{L}_{\Delta}(\mathcal{N}^v_{deact}) \gg \sum_{d=1}^{D} \mathcal{L}_{\Delta}(\mathcal{N}^{rand}_{deact}),
\end{equation}
where $\mathcal{L}_{\Delta}$ represents the degradation in generation quality (e.g., increase in loss or decrease in clinical accuracy metrics) when neurons are deactivated.

\noindent\textbf{View-specific Neuron Fine-tuning (VNF).}
Once identified and verified, we employ a targeted fine-tuning mechanism. Unlike standard fine-tuning methods which updates all parameters, VNF freezes the majority of the model and optimizes \textit{only} the identified pattern neurons $\mathcal{N}^v$. This allows the model to enhance its representation of specific views without catastrophically forgetting general language generation capabilities. The optimization objective is the standard cross-entropy loss over the view-specific fine-tuning set $\mathcal{D}_{ft}$:
\begin{equation}
\mathcal{L}_{VNF} = -\sum_{(x, y) \in \mathcal{D}_{ft}} \sum_{t=1}^{T} \log P(y_t | y_{<t}, x; \theta_{\mathcal{N}^v}),
\end{equation}
where $\theta_{\mathcal{N}^v}$ denotes that only the parameters of the pattern neurons are updated.

\begin{table}[ht]
\footnotesize
\setlength{\fboxsep}{0.5pt}
\setlength{\tabcolsep}{9pt}
\renewcommand{\arraystretch}{0.4}
\begin{center}
\caption{NLG performance comparison of the integrated reports and reports generated by existing SOTA RRG methods on IU-XRay dataset. The subscripts MedGemma, InternVL, LLaVA, and Hulu denote that the VLM backbones used are MedGemma-4B, InternVL-DA-Medical-4B, LLaVA-Med-7B, and Hulu-Med-7B, respectively. BL: BLEU, MTR: METEOR, RG-L: ROUGE-L. Higher value indicates better performance.}
\label{tab:iuxray_results}
\begin{tabular}{@{}l |c c c c c c @{}}
\toprule
\textsc{\textbf{Method}} & \textbf{BL-1} & \textbf{BL-2} & \textbf{BL-3} & \textbf{BL-4} & \textbf{MTR}  & \textbf{RG-L} \\
\midrule 
R2GEN~\cite{chen2020generating} & $0.390$ & $0.293$ & $0.236$ & $0.161$ & $0.187$ & $0.371$ \\
R2GENCMN~\cite{chen2022cross} & $0.402$ & $0.310$ & $0.225$ & $0.175$ & $0.190$ & $0.365$ \\
XPRONet~\cite{wang2022cross} & $0.480$ & $0.312$ & $0.225$ & $0.175$ & $0.190$ & $0.364$ \\
RGRG~\cite{tanida2023interactive} & $0.373$ & $0.249$ & $0.175$ & $0.126$ & $0.168$ & $0.264$ \\
M2KT~\cite{yang2023radiology} & $0.489$ & $0.315$ & $0.228$ & $0.174$ & -- & $0.379$ \\
CAMANET~\cite{wang2024camanet} & $0.478$ & $0.314$ & $0.229$ & $0.179$ & $0.202$ & $0.361$ \\
PromptMRG~\cite{jin2024promptmrg} & $0.488$ & $0.253$ & $0.237$ & $0.189$ & $0.192$ & $0.374$ \\
\cmidrule{1-7}
View-PNDF$_{\mathrm{MedGemma}}$ & $0.496$ & $0.322$ & $0.240$ & $0.194$ & $0.205$ & $0.385$ \\
View-PNDF$_{\mathrm{InternVL}}$& $0.492$ & $0.318$ & $0.234$ & $0.191$ & $0.198$ & $0.382$ \\
View-PNDF$_{\mathrm{LLaVA}}$& $0.505$ & $0.330$ & $0.248$ & $0.201$ & $0.212$ & $0.392$ \\
View-PNDF$_{\mathrm{Hulu}}$ & $0.510$ & $0.335$ & $0.252$ & $0.205$ & $0.216$ & $0.396$ \\
\cmidrule{1-7}
\textbf{Average} & $\underline{\textbf{0.501}}$ & $\underline{\textbf{0.326}}$ & $\underline{\textbf{0.244}}$ & $\underline{\textbf{0.198}}$ & $\underline{\textbf{0.208}}$ & $\underline{\textbf{0.389}}$ \\
\bottomrule
\end{tabular}
\end{center}
\end{table}

\begin{table}[ht]
\footnotesize
\setlength{\fboxsep}{0.5pt}
\setlength{\tabcolsep}{2.0pt}
\renewcommand{\arraystretch}{0.5}
\begin{center}
\caption{NLG and CE performance comparison of the integrated reports and reports generated by existing SOTA methods on MIMIC-CXR dataset. The subscripts MedGemma, InternVL, LLaVA, and Hulu denote that the VLM backbones used are MedGemma-4B, InternVL-DA-Medical-4B, LLaVA-Med-7B, and Hulu-Med-7B, respectively. BL: BLEU, MTR: METEOR, RG-L: ROUGE-L, CE: Clinical Efficacy, Prec.: Precision, Rec.: Recall.}
\label{tab:mimic_results}
\begin{tabular}{@{}l|c c c c c c c c c @{}}
\toprule
\multirow{2}{*}{\textsc{\textbf{Method}}} & \multicolumn{6}{c}{\textbf{NLG Metrics}} & \multicolumn{3}{c}{\textbf{CE Metrics}} \\
\cmidrule(lr){2-7} \cmidrule(lr){8-10}
& \textbf{BL-1} & \textbf{BL-2} & \textbf{BL-3} & \textbf{BL-4} & \textbf{MTR}  & \textbf{RG-L} & \textbf{Prec.} & \textbf{Rec.} & \textbf{F1} \\
\midrule 
R2GEN~\cite{chen2020generating} & $0.353$ & $0.218$ & $0.145$ & $0.103$ & $0.142$ & $0.270$ & $0.406$ & $0.203$ & $0.204$ \\
R2GENCMN~\cite{chen2022cross} & $0.348$ & $0.206$ & $0.135$ & $0.094$ & $0.136$ & $0.266$ & $0.440$ & $0.325$ & $0.374$ \\
XPRONet~\cite{wang2022cross} & $0.344$ & $0.215$ & $0.146$ & $0.105$ & $0.138$ & $0.279$ & $0.463$ & $0.285$ & $0.353$ \\
M2KT~\cite{yang2023radiology} & $0.368$ & $0.220$ & $0.149$ & $0.107$ & -- & $0.259$ & $0.420$ & $0.339$ & $0.352$ \\
CAMANET~\cite{wang2024camanet} & $0.367$ & $0.219$ & $0.151$ & $0.107$ & $0.142$ & $0.279$ & $0.467$ & $0.323$ & $0.381$ \\
PromptMRG~\cite{jin2024promptmrg} & $0.369$ & $0.223$ & $0.142$ & $0.099$ & $0.147$ & $0.260$ & $0.471$ & $0.198$ & $0.352$ \\
MLRG~\cite{liu2025enhanced} & $0.398$ & $0.236$ & $0.159$ & $0.156$ & $0.169$ & $0.278$ & $0.491$ & $0.398$ & $0.401$ \\
\cmidrule{1-10}
View-PNDF$_{\mathrm{MedGemma}}$ & $0.408$ & $0.245$ & $0.166$ & $0.164$ & $0.174$ & $0.286$ & $0.502$ & $0.415$ & $0.408$ \\
View-PNDF$_{\mathrm{InternVL}}$  & $0.402$ & $0.240$ & $0.162$ & $0.160$ & $0.168$ & $0.282$ & $0.495$ & $0.405$ & $0.398$ \\
View-PNDF$_{\mathrm{LLaVA}}$ & $0.418$ & $0.254$ & $0.174$ & $0.171$ & $0.182$ & $0.294$ & $0.515$ & $0.432$ & $0.420$ \\
View-PNDF$_{\mathrm{Hulu}}$ & $0.425$ & $0.260$ & $0.179$ & $0.176$ & $0.188$ & $0.301$ & $0.522$ & $0.445$ & $0.428$ \\
\cmidrule{1-10}
\textbf{Average} & $\underline{\textbf{0.413}}$ & $\underline{\textbf{0.250}}$ & $\underline{\textbf{0.170}}$ & $\underline{\textbf{0.168}}$ & $\underline{\textbf{0.178}}$ & $\underline{\textbf{0.291}}$ & $\underline{\textbf{0.509}}$ & $\underline{\textbf{0.424}}$ & $\underline{\textbf{0.414}}$ \\
\bottomrule
\end{tabular}
\end{center}
\end{table}

\section{Experiments and Results}

\subsection{Dataset and Evaluation Metrics}

\noindent\textbf{Datasets.} 
We evaluate on \textbf{IU-XRay}~\cite{demner2015preparing} and \textbf{MIMIC-CXR}~\cite{johnson2019mimic}, both containing frontal and lateral views. For IU-XRay ($2,955$ samples), we adopt the standard 70\%/10\%/20\% 
split~\cite{jing2020show}. For MIMIC-CXR, to ensure balanced view representation, we curate a subset of 
$10,000$ training samples and $1,000$ validation/test samples, instead of the standard split~\cite{chen2020generating}, which is dominated by single-view reports.
\newline
\textbf{Evaluation Metrics.} 
Integrated reports are evaluated using NLG metrics, including 
BLEU~\cite{papineni2002bleu}, METEOR~\cite{denkowski2011meteor}, and  ROUGE-L~\cite{lin2004rouge}. For view-specific reports, given the limited ability of n-gram metrics to assess 
clinical validity, we employ GPT-4o~\cite{achiam2023gpt} to score semantic fidelity on a scale of 1--10, following recent LLM-based evaluation practices~\cite{zheng2023judging}.


\begin{table}[t]
\centering
\captionsetup{justification=centering,singlelinecheck=false}
\footnotesize

\savebox{\topbox}{%
  \renewcommand{\arraystretch}{0.8}%
  \setlength{\tabcolsep}{6pt}%
  \begin{tabular}{@{}l cccc cccc@{}}
    \toprule
    \multirow{3}{*}{\textbf{Model}}
      & \multicolumn{4}{c}{\textbf{IU-XRay~\cite{demner2015preparing}}}
      & \multicolumn{4}{c}{\textbf{MIMIC-CXR~\cite{johnson2019mimic}}} \\
    \cmidrule(lr){2-5}\cmidrule(lr){6-9}
      & \multicolumn{2}{c}{\textbf{Frontal}} & \multicolumn{2}{c}{\textbf{Lateral}}
      & \multicolumn{2}{c}{\textbf{Frontal}} & \multicolumn{2}{c}{\textbf{Lateral}} \\
    \cmidrule(lr){2-3}\cmidrule(lr){4-5}\cmidrule(lr){6-7}\cmidrule(lr){8-9}
      & \textbf{w/o} & \textbf{w/} & \textbf{w/o} & \textbf{w/}
      & \textbf{w/o} & \textbf{w/} & \textbf{w/o} & \textbf{w/} \\
    \midrule
    MedGemma-4B~\cite{sellergren2025medgemma}     
      & 7.1 & 7.3 & 2.9 & 6.8 & 6.2 & 6.3 & 3.6 & 5.9 \\
    InternVL-DA-Medical-4B~\cite{chen2024internvl} 
      & 8.5 & 8.6 & 3.7 & 7.9 & 7.4 & 7.6 & 2.9 & 7.1 \\
    LLaVA-Med-7B~\cite{li2023llava}                
      & 8.7 & 8.8 & 4.1 & 8.3 & 8.3 & 8.5 & 3.9 & 7.8 \\
    Hulu-Med-7B~\cite{jiang2025hulu}               
      & 8.9 & 9.1 & 4.3 & 8.5 & 8.4 & 8.5 & 4.2 & 8.1 \\
    \bottomrule
  \end{tabular}%
}

\begin{minipage}{\wd\topbox}
  \centering
  \caption{Performance comparison of medical VLMs with and without View-PNDF. Scores are evaluated by GPT-4o.}
  \label{tab:comparison}
  \usebox{\topbox}
\end{minipage}

\begin{minipage}{\wd\topbox}
\centering

\begin{minipage}[t]{0.52\linewidth}
  \centering
  \caption{NLG results of the InternVL-DA-Medical-4B on the IU-Xray dataset.}
  \label{FT}

  \renewcommand{\arraystretch}{0.8}
  \setlength{\tabcolsep}{4pt}

  \begin{tabular}{@{}lcccc@{}}
    \toprule
    \textbf{Before FT} & \textbf{BL-1} & \textbf{BL-4} & \textbf{MTR} & \textbf{RG-L} \\
    \midrule
    Frontal & 0.398 & 0.156 & 0.163 & 0.265 \\
    Lateral & 0.397 & 0.154 & 0.162 & 0.266 \\
    \midrule
    \textbf{After FT} & \textbf{BL-1} & \textbf{BL-4} & \textbf{MTR} & \textbf{RG-L} \\
    \midrule
    Frontal & 0.397 & 0.155 & 0.164 & 0.265 \\
    Lateral & 0.397 & 0.154 & 0.163 & 0.264 \\
    \bottomrule
  \end{tabular}
\end{minipage}
\hfill
\begin{minipage}[t]{0.42\linewidth}
  \centering
  \caption{LLM evaluation scores of InternVL-DA-Medical-4B on the IU-Xray dataset.}
  \label{tab:eval}

  \renewcommand{\arraystretch}{1.42}
  \setlength{\tabcolsep}{4pt}

  \begin{tabular}{@{}lcc@{}}
    \toprule
    \textbf{Evaluator} & \textbf{Frontal} & \textbf{Lateral} \\
    \midrule
    GPT-4o      & 8.7 & 7.5 \\
    Qwen-72B    & 8.7 & 7.7 \\
    DeepSeek-V3 & 8.6 & 7.6 \\
    \bottomrule
  \end{tabular}
\end{minipage}

\end{minipage}

\end{table}

\begin{figure}[t]
\centering
  \includegraphics[width=1.0\linewidth, height=0.5\textheight, keepaspectratio]{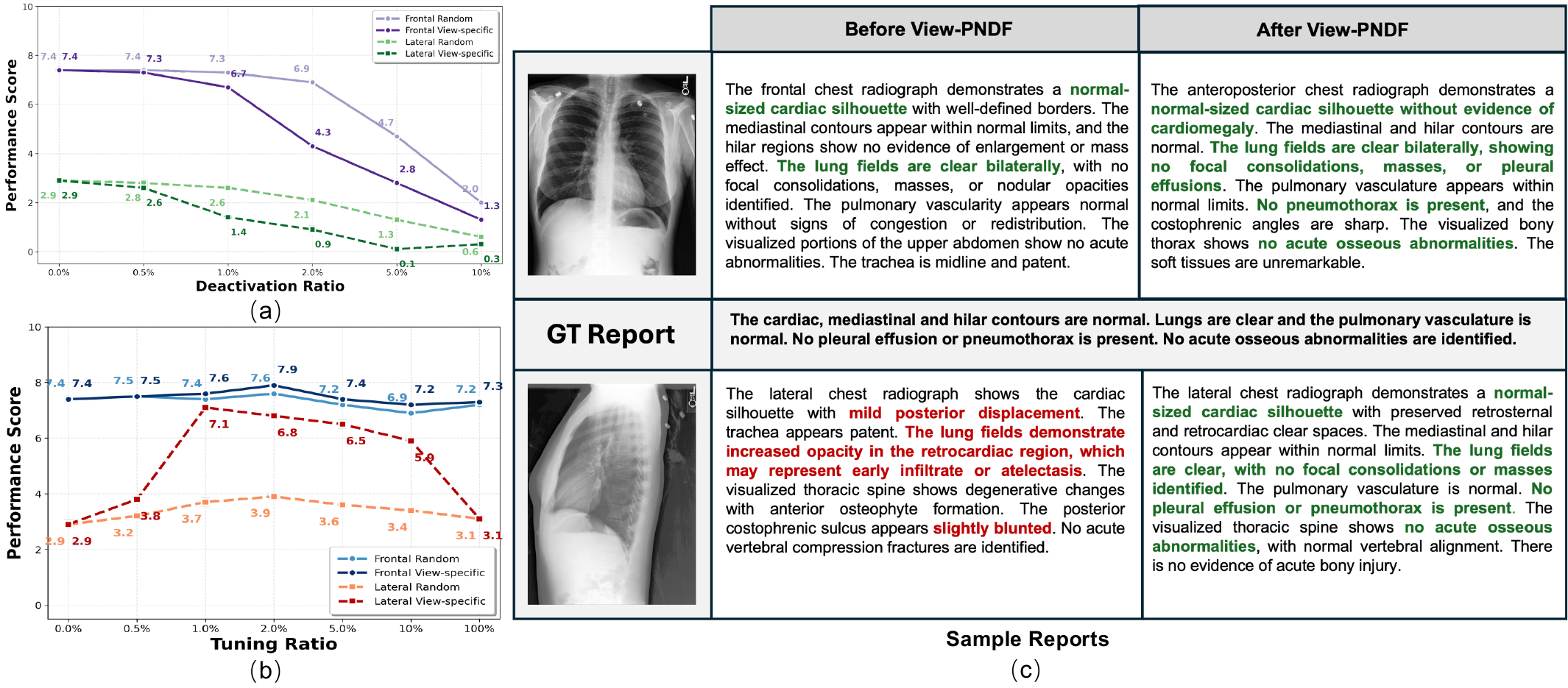}
\caption{Ablation studies of VND, VNV, and VNF on MIMIC-CXR dataset. (a) VND yields steeper performance drops than random deactivation, confirming neuron specificity. (b) VNF boosts lateral metrics while maintaining frontal quality. (c) Sample output from InternVL-DA-Medical-4B.}
\label{fig:frontal} 
\end{figure}

\subsection{Implementation}
The model is trained using a PyTorch implementation on two NVIDIA A6000 GPUs. 
The initial learning rate is set as $4\times 10^{-6}$ with a cosine learning rate schedule. We use AdamW~\cite{loshchilov2017decoupled} optimizer to optimize the network with a weight decay of $0.5$. We train the model for 1 epoch with a batch size of 8. The warm up ratio is set as $0.3$.
\subsection{Comparative Experiment}
We evaluated our method on IU-XRay~\cite{demner2015preparing} and MIMIC-CXR~\cite{johnson2019mimic} against state-of-the-art methods. As shown in Tab.~\ref{tab:iuxray_results}, applying our method to four VLM backbones (MedGemma, InternVL, LLaVA, Hulu) on IU-XRay consistently outperformed baselines across all metrics, with View-PNDF$_{\mathrm{Hulu}}$ achieving the best results (e.g., $0.510$ BLEU-1). On the MIMIC-CXR dataset (Tab.~\ref{tab:mimic_results}), our approach maintained this advantage, surpassing the leading baseline MLRG in both linguistic quality (ROUGE-L $0.291$ vs $0.278$) and Clinical Efficacy (F1 $0.414$ vs $0.401$). These results confirm that fine-tuning view-specific neurons significantly enhances both the linguistic fluency and clinical accuracy of generated reports.

\subsection{Ablation Study}
To verify the effectiveness of each module, we conduct ablation studies on the IU-XRay~\cite{demner2015preparing} and MIMIC-CXR~\cite{johnson2019mimic} datasets using the InternVL medical model~\cite{chen2024internvl}. 
\newline

\noindent\textbf{The Effectiveness of View-PNDF.} Tab.~\ref{tab:comparison} details the GPT-4o evaluation of four VLMs with and without View-PNDF. The method consistently improved performance across all models and datasets. Notably, lateral view scores saw significant gains (e.g., +4.2 points on MIMIC-CXR for InternVL), demonstrating that View-PNDF effectively captures view-specific patterns to generate more accurate descriptions for both views.

\noindent\textbf{The Effectiveness of VND and VNV.} To validate our detection mechanism, we compared view-specific versus random neuron deactivation (Fig.~\ref{fig:frontal} (a)). View-specific deactivation caused drastic performance drops (e.g., lateral score dropping from 2.9 to 0.3 on MIMIC-CXR), whereas random deactivation had minimal impact. This significant disparity confirms that VND successfully identifies neurons critical for specific views.

\noindent\textbf{The Effectiveness of VNF.} Fig.~\ref{fig:frontal} (b) shows that view-specific fine-tuning significantly boosts lateral view performance (e.g., reaching 7.3 vs. 3.7 for random tuning on MIMIC-CXR) while preserving or enhancing frontal view quality. This proves VNF effectively bridges the performance gap between views without compromising general capabilities.

\noindent\textbf{LLM Evaluators.}
Tab.~\ref{FT} shows that standard NLG metrics yield nearly identical scores before and after fine-tuning, failing to capture clinical nuances. In contrast, evaluations using GPT-4o~\cite{achiam2023gpt}, Qwen~\cite{bai2023qwen}, and DeepSeek~\cite{bi2024deepseek} (Tab.~\ref{tab:eval}) reveal clear performance differences, demonstrating that LLM-based evaluation aligns better with clinical validity for this task.

\noindent\textbf{Qualitative Results.}
Fig.~\ref{fig:frontal} (c) illustrates the qualitative impact of View-PNDF. While the baseline model misses critical findings (e.g., sternal wires, cardiomegaly) and lacks inter-view consistency, the fine-tuned model successfully captures these pathologies across both frontal and lateral views. This confirms that our method enhances the detection of view-specific pathological patterns, yielding more comprehensive reports.

\section{Conclusion}
We proposed View-PNDF to detect and fine-tune view-specific pattern neurons in RRG models. Experiments confirm the existence of these neurons and show that optimizing a small fraction ($<5.0\%$) significantly enhances generation consistency and accuracy across views. While demonstrated on chest X-rays, our parameter-efficient method holds potential for other modalities (e.g., CT, MRI). Future work will focus on extending this approach to multi-modal data and developing model-agnostic implementations.

\begin{credits}
\subsubsection{\ackname}
This project is supported by the Ministry of Education, Singapore, under its Academic Research Fund Tier 1 RS16/23. This Project is also partly supported by Centre of AI in Medicine, Singapore.

\subsubsection{\discintname}
The authors have no competing interests to declare that are relevant to the content of this article.
\end{credits}

%
%
%
\bibliographystyle{splncs04}
\bibliography{ref}
\end{document}